\documentclass[a4paper,twoside]{article}

\usepackage{epsfig}
\usepackage{subcaption}
\usepackage{calc}
\usepackage{amssymb}
\usepackage{amstext}
\usepackage{amsmath}
\usepackage{amsthm}
\usepackage{multicol}
\usepackage{pslatex}
\usepackage{apalike}
\usepackage{orcidlink}
\usepackage{algorithm2e}
\usepackage[bottom]{footmisc}
\usepackage{SCITEPRESS}
\usepackage[capitalise]{cleveref}
\creflabelformat{equation}{#2#1#3}
\usepackage{cite}

\def\BibTeX{{\rm B\kern-.05em{\sc i\kern-.025em b}\kern-.08em
T\kern-.1667em\lower.7ex\hbox{E}\kern-.125emX}}

\begin{document}

\title{Optimization of Link Configuration for Satellite Communication Using Reinforcement Learning}

\author{Anonymous Authors}
\author{\authorname{Tobias Rohe\sup{1}\orcidlink{0009-0003-3283-0586}, Michael Kölle\sup{1}\orcidlink{0000-0002-8472-9944}, Jan Matheis\sup{2}, Rüdiger Höpfl\sup{2}, Leo Sünkel\sup{1}\orcidlink{0009-0001-3338-7681} and Claudia Linnhoff-Popien\sup{1}\orcidlink{0000-0001-6284-9286}}
\affiliation{\sup{1}Mobile and Distributed Systems Group, LMU Munich, Germany}
\affiliation{\sup{2}Airbus Defence and Space, Germany}
\email{tobias.rohe@ifi.lmu.de}
}


\abstract{
Satellite communication is a key technology in our modern connected world. With increasingly complex hardware, one challenge is to efficiently configure links (connections) on a satellite transponder. Planning an optimal link configuration is extremely complex and depends on many parameters and metrics. The optimal use of the limited resources, bandwidth and power of the transponder is crucial. Such an optimization problem can be approximated using metaheuristic methods such as simulated annealing, but recent research results also show that reinforcement learning can achieve comparable or even better performance in optimization methods. However, there have not yet been any studies on link configuration on satellite transponders. In order to close this research gap, a transponder environment was developed as part of this work. For this environment, the performance of the reinforcement learning algorithm PPO was compared with the metaheuristic simulated annealing in two experiments. The results show that Simulated Annealing delivers better results for this static problem than the PPO algorithm, however, the research in turn also underlines the potential of reinforcement learning for optimization problems.
}

\onecolumn \maketitle \normalsize \setcounter{footnote}{0} \vfill

\section{INTRODUCTION} \label{sec:introduction}
In today’s world, satellite communication is a pivotal technology, utilized across a broad spectrum of modern applications. While terrestrial communication systems depend on conventional infrastructure that is often vulnerable to disruptions, satellite communication has consistently demonstrated high reliability. This reliability becomes especially crucial in scenarios such as natural disasters, where large-scale power outages often occur. To further enhance the reliability and efficiency of satellite communication, reducing disruption, and optimize bandwidth usage, artificial intelligence (AI) is increasingly being employed~\cite{vazquez2020use}.

A critical factor in the performance of satellite communication networks is determining the optimal link configuration of the satellite transponder — an inherently complex optimization problem. The optimal configuration of these links changes dynamically as links are terminated or added. This dynamic nature introduces the potential of reinforcement learning (RL) as a method to optimize the link configuration.

By leveraging learned metrics and the ability to generalize, a RL approach could offer faster and superior outcomes compared to traditional metaheuristic approaches. In this paper, we model a satellite transponder environment and illustrate how a meaningful action and observation space can be defined for this purpose. Since modeling the dynamic aspects of this problem exceeds the scope of this work, we have chosen a static approach. In this context, static refers to finding the best possible link configuration for a restricted number of links with predetermined parameter configurations. Building upon previous research, where RL has achieved equal or better results than metaheuristics for static optimization problems~\cite{klar2023performance}, we extend this investigation to the specific optimization problem of link configuration in satellite transponders.

The remainder of the paper is structured as follows: Section~\ref{sec:preliminaries} provides an overview of the background relevant to the study, followed by a discussion of related work in Section~\ref{sec:related-work}. In Section~\ref{sec:approach}, we present our proposed approach for optimizing the link configuration. The experimental setup is detailed in Section~\ref{sec:experimental-setup}, and the results of the experiments are discussed in Section~\ref{sec:results}. Finally, Section~\ref{sec:conclusion} concludes the paper and outlines potential directions for future research.

\section{BACKGROUND} \label{sec:preliminaries}
\subsection{Reinforcement Learning}
Machine learning is a subfield of artificial intelligence, which can be further broken down into three categories: supervised learning, unsupervised learning, and reinforcement learning (RL). RL is particularly well-suited for addressing sequential decision-making problems, such as those encountered in chess or in the game AlphaGo~\cite{deliu2023reinforcement}, where significant successes were achieved in recent years~\cite{silver2018general}. In RL, an agent learns a policy $\pi$ by interacting with an environment through trial and error, with the goal of making optimal decisions. The two main components in RL are the environment and the agent. The environment is often modeled as a simulation of the real world, as an agent interacting directly with the real environment may be infeasible, too risky or too expensive~\cite{prudencio2023survey}.

The foundation of RL lies in the Markov Decision Process (MDP). In an MDP, the various states of the environment are represented by states in the process. At any given time $t$, the agent occupies a state $S_t$. From this state, the agent can select from various actions to transition to other states. Thus, the system consists of state-action pairs, or tuples $(A_t, S_t)$. When the agent selects an action $A_t$ and executes it in the environment, the agent receives feedback. This feedback includes an evaluation of the performed action in the form of a reward, and the new state $S_{t+1}$ of the environment. The reward may be positive or negative, representing either a benefit or a penalty. The agent's objective is to maximize the accumulated reward, as shown in Equation~\ref{eq:optimal_policy}. 

\begin{equation}
\pi^* = \arg\max_{\pi} \mathbb{E} \left[ \sum_{t=0}^{H} \gamma^t \cdot R(s_t, a_t) \right]    
\label{eq:optimal_policy}
\end{equation}

A major challenge in finding the optimal policy $\pi^*$ is the balance between exploration and exploitation. Exploration refers to the degree to which the agent explores the environment for unknown states, while exploitation refers to the degree to which the agent applies its learned knowledge to achieve the highest possible reward. At the start of training, the agent should focus more on exploring the environment, even if this means not always selecting the action that yields the highest immediate reward (i.e., avoiding a greedy strategy). This approach allows the agent to gain a more comprehensive understanding of the environment and, ultimately, develop a better policy. As the learning process progresses, the agent should shift towards exploiting its knowledge more and exploring less.

In RL, there is a distinction between model-based and model-free approaches. In model-free RL, the agent directly interacts with the environment and receives feedback. The environment may be a real-world environment or a simulation. In contrast, in model-based RL, the agent also interacts with the environment, but it can update its policy using a model of the environment. In addition to standard RL, there exists an advanced method called deep reinforcement learning (deep RL). In deep RL, neural networks are used to approximate the value function $\hat{v}(s; \theta)$ or $\hat{q}(s, a;\theta)$ of the policy $\pi(a \mid s; \theta)$ or the model. The parameter $\theta$ represents the weights of the neural network. The use of neural networks enables the learning of complex tasks. During training, the connections between neurons, also called nodes, are dynamically adjusted (weighted) to improve the quality of the function approximation.

\section{RELATED WORK} \label{sec:related-work}
Metaheuristics are commonly applied to optimization problems and frequently yield efficient near-optimal solutions in complex environments. A promising alternative to these traditional approaches are RL-algorithms, whose potential in the context of optimization problems has already been scientifically explored~\cite{zhang2022deep, ardon2022reinforcement, li2021deep, kolle2024study}. Studies such as~\cite{klar2023performance} and~\cite{mazyavkina2021reinforcement} demonstrate that RL-algorithms can achieve results similar to or even superior to those obtained by standard metaheuristics. This applies to a variety of optimization problems, including the Traveling Salesman Problem (TSP)~\cite{bello2016neural}, the Maximum Cut (Max-Cut) problem, the Minimum Vertex Cover (MVC) problem, and the Bin Packing Problem (BPP)~\cite{mazyavkina2021reinforcement}.

While the work by~\cite{klar2023performance} does not focus on the arrangement of links on a transponder, it focuses on the planning of factory layouts, which is also a static optimization problem. This study compares a RL-based approach with a metaheuristic approach and examines which method proves more effective. The research highlights RL's capability to learn problem-specific application scenarios, making it an intriguing alternative to conventional methods. Given the uniqueness of each problem and the limited amount of scientific research on this specific topic, individual consideration and analysis of each problem type are essential, which is also goal of this research. 
\section{APPROACH} \label{sec:approach}
\subsection{System Description}
\begin{figure}[h!]
    \centering
    \includegraphics[width=\linewidth]{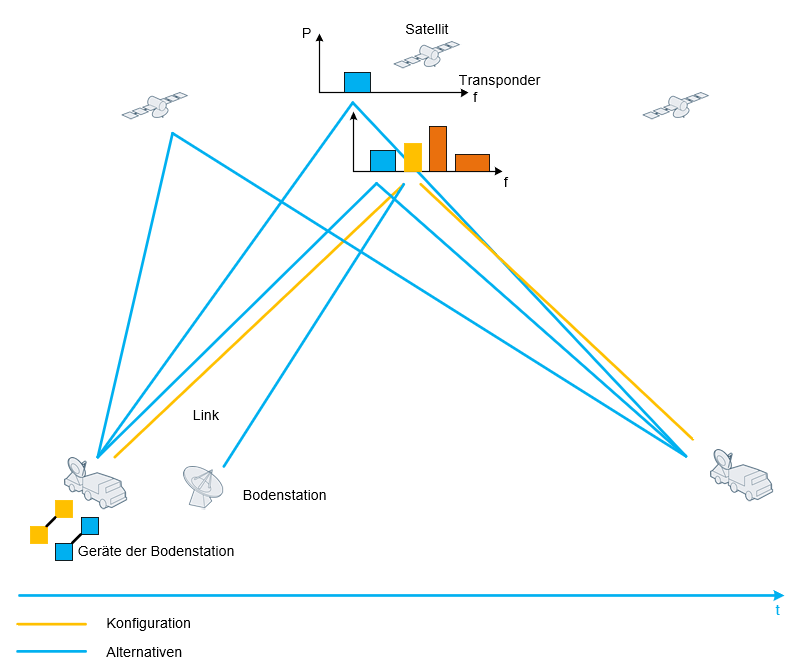}
    \caption{Illustration of Satellite Communication}
    \label{fig:communication_scheme}
\end{figure}

In satellite communication, ground stations are responsible for sending and receiving data. The data is transmitted from the sending ground station to the satellite via communication links, also referred to as links or carriers, and then relayed from the satellite to the receiving ground station via another link. On the satellite itself, the links pass through chains of devices known as transponders. Incoming signals are received by an antenna on the satellite, amplified, converted in frequency, and retransmitted through the antenna. A transponder is a component within a satellite that receives signals transmitted from a ground station, amplifies them with a specific amplification factor using the available power, changes their frequency to avoid interference, and then retransmits the signals back to Earth. This process enables the efficient transmission of information across vast distances in space. Figure \ref{fig:communication_scheme} illustrates the path of the links as they move from one ground station through the satellite’s transponder to another ground station, facilitating data transmission.

However, each transponder has limited resources in terms of bandwidth and power. It is crucial that the links are configured to consume as few resources as possible. A link is characterized by several parameters, such as data rate, MOD-FEC combination (Modulation and Forward Error Correction), bandwidth, and EIRP (Effective Isotropic Radiated Power). To ensure the data is transmitted with sufficient quality, a minimum performance level must be maintained for each link. The challenge lies in selecting the optimal parameters for each link so that they use the least possible resources on the transponder while maintaining the required communication quality. This task is further complicated because the bandwidth is a function of various other parameters. In the scenario considered here, a fixed number of links are established through which data is transmitted. Once the transmission is complete, all links are simultaneously taken down.

\subsection{Systemparameters}
The following section outlines the process of establishing a satellite connection. To set up a link from a ground station to a satellite, various parameters must be carefully combined. The choice of these parameters determines how efficiently the links can be arranged on the satellite transponder, minimizing the consumption of transponder resources, specifically EIRP and bandwidth.

One key parameter is the EIRP, which represents the power used to establish the link. Each link requires a certain EIRP to ensure that the transmitted data arrives with the desired quality. The necessary EIRP depends on the data rate to be transmitted. The data rate, typically measured in bits per second (bps), indicates how much data can be transmitted per unit of time and plays a crucial role in determining the required bandwidth.

The MOD-FEC combination is another essential parameter. Modulation (MOD) is the process of converting data into an electrical signal (electromagnetic wave) for transmission, where characteristics of the wave—such as amplitude, frequency, or phase—are adjusted to encode the data. Forward Error Correction (FEC) enables error detection and correction without the need for retransmission, by adding correction information. The combination of modulation and FEC greatly influences transmission efficiency, especially in environments where interference is likely. Various valid MOD-FEC combinations are available for the transmission of a link, typically determined by the hardware in use.

Another important parameter is bandwidth, which describes the ability of a communication channel or transmission medium to transmit data over a specified time period. Measured in Hertz (Hz), bandwidth directly influences transmission speed. In satellite communication, bandwidth refers to the frequency range over which a signal is transmitted. Both the data rate and the MOD-FEC combination influence the determination of the bandwidth. To compute the bandwidth, several parameters are required: FEC, MOD, OH factor, RS factor, overhead, spacing factor, rollout factor, and the data rate. To simplify the model of the environment, fixed values were assigned to the OH factor, RS factor, overhead, spacing factor, and rollout factor. These parameters do not play a role in the subsequent experiment but were included in the calculation for the sake of completeness.

Calculating the bandwidth of a link involves three steps. First, the transmission rate \text{\"UR} is calculated by multiplying the parameters data rate, OH factor, RS factor, and FEC. This is shown in Equation \ref{eq:ur_calc}.

\begin{equation}
    \text{\"UR} = \text{data rate} \ast \text{OHFactor} \ast \text{RSFactor} \ast \text{FEC}
    \label{eq:ur_calc}
\end{equation}

The transmission rate is then multiplied by the MOD, resulting in the symbol rate (SR).

\begin{equation}
    \text{SR} = \text{\"UR} \ast \text{MOD}
    \label{eq:sr_calc}
\end{equation}

Finally, the SR is multiplied by the Rollout Factor and the Spacing Factor to obtain the bandwidth of a link.

\begin{equation}
    \begin{split}
        \text{Bandwidth} = \text{SR} \ast \text{Rollout Factor} \ast \text{Spacing Factor}
    \end{split}
    \label{eq:bandwith_calc}
\end{equation}

The last important parameter is the center frequency, which determines the position of the links on the transponder.

Link configuration is usually carried out manually with the support of software applications. However, this is not an optimal procedure, as the ideal solution is rarely found and therefore does not save resources. In addition, the link configuration has to be calculated manually each time. 

\subsection{Action and Observation Space} \label{sec:Action_and_Observation_Space}

In RL, the action space represents the set of all possible actions that an agent can perform within an environment. The agent responds to received observations by choosing an action. To analyze how different action spaces affect performance, two distinct action spaces were defined.

The first action space (Action Space 1) was defined using a nested dictionary from the Gymnasium framework (gymnasium.spaces.Dict). This dictionary contains three sub-dictionaries, each representing a link. Each link dictionary includes the parameters MOD-FEC combination, center frequency, and EIRP. Both EIRP and center frequency are continuous values, so a float value between 0 and 1 is selected at each step within the environment. The MOD-FEC combination, however, is a discrete value. A distinctive feature of Action Space 1 is that all parameters can be selected for any link at each step, allowing the parameters to be reset at every action.

The second action space (Action Space 2) adopts a different approach. Similar to Action Space 1, the number of links is fixed at exactly three. However, the action space is represented as a 4-tuple (link, parameter, continuous, discrete). In this setup, a specific parameter is chosen for a link, and the value of that parameter is modified. As with Action Space 1, EIRP and center frequency are continuous values, while the MOD-FEC combination is represented by discrete values.

The observation space in RL represents the set of all possible states or information transmitted from the environment to the agent. It contains all the necessary information to reconstruct the current state of the environment, including details about the link configurations on the transponder. For each link, the observation space includes the parameters: center frequency, EIRP, MOD-FEC combination, and Link Bandwidth Percentage. The Link Bandwidth Percentage represents the portion of the transponder’s bandwidth allocated to a particular link. This is calculated by dividing the bandwidth of the link by the total transponder bandwidth. The observation space remains consistent for both Action Space 1 and Action Space 2.

\subsection{Simplifications}
In the experiments conducted, the task was simplified in several aspects to facilitate the learning process. Specifically, the transponder environment was simplified. For instance, the possible MOD-FEC combinations were restricted, and the modeling of the links in the frequency domain was simplified, excluding the interference of electromagnetic waves (links) which normally affect communication quality.

Additionally, simplifications were made regarding the inherently dynamic nature of the problem. Currently, a Proximal Policy Optimization (PPO) model can only be trained for a fixed number of links defined prior to training, which represents a static rather than dynamic scenario. In our experiments, a PPO model was trained exclusively for three links, and dynamic changes — such as the addition or termination of links — were not accounted for. The current environment model is configured so that each link can be assigned one of three possible MOD-FEC combinations, and the data rate for each link is set to a uniform value.

Future iterations of the model will progressively address these simplifications to handle the more complex tasks encountered in real-world scenarios.
\section{EXPERIMENTAL SETUP} \label{sec:experimental-setup}

Two experiments were conducted, differing only in the configuration of the RL model. The PPO algorithm was utilized in both models, with the only distinction being the action space, as explained in Chapter~\ref{sec:Action_and_Observation_Space}. For both experiments, Python (version 3.8.10), PyTorch (version 1.13.1), Gymnasium (version 0.28.1), and Ray RLlib (version 0.0.1) were used.

\subsection{PPO Algorithm}
The PPO model serves as the baseline against which the Random Action and Simulated Annealing baselines are compared in terms of performance. A neural network was used to implement the PPO, tasked with calculating the policy parameters. The neural network topology consisted of the observation space as the input layer, followed by two hidden layers with 256 neurons each, and the action space as the output layer. The learning rate for the PPO model was optimized, with further details provided in the section "Training Process". The hyperparameters are shown in Table \ref{tab:ppo_hyperparams}.

\begin{table}[h!]
\centering
\begin{tabular}{|l|l|l|}
\hline
\textbf{Parameter} & \textbf{Datatype} & \textbf{Value} \\ \hline
Discount Factor    & Float             & 0.99          \\ \hline
Batch Size         & Integer           & 2000          \\ \hline
Clipping Range     & Float             & 0.3           \\ \hline
GAE Lambda         & Float             & 1.0           \\ \hline
Entropy coeff      & Float             & 0.0           \\ \hline
Number of SGD      & Integer           & 30            \\ \hline
Minibatchsize      & Integer           & 128           \\ \hline
\end{tabular}
\caption{PPO Algorithm Hyperparameter Configuration}
\label{tab:ppo_hyperparams}
\end{table}

\subsection{Random Action Strategy}
The naïve random action strategy, which is compared with the PPO, consists of choosing actions purely at random. We define a fundamental learning success here if the PPO model achieves better results than the random action model.

\subsection{Simulated Annealing}
Simulated annealing follows a structured process: Initially, the algorithm is provided with a randomly generated action. This action is evaluated by the cost function, which in this case is represented by the step function of the environment. The corresponding reward for the action is calculated. Next, the neighbor function generates a new action. If the reward of the new action is better than that of the previous action, it is accepted as the new action. However, if the reward is lower, the new action is not immediately discarded. An additional mechanism, known as temperature, plays a key role in determining the probability of selecting a worse action as the new action.

At the beginning of the algorithm, the temperature is high, leading to a higher probability of choosing a worse action. This facilitates broader exploration in the search for the global optimum. As the algorithm progresses, the temperature decreases gradually. As the temperature lowers, the probability of accepting a worse action diminishes, leading to a more focused search for the local optimum and enabling fine-tuning of the solution.

\begin{table}[h!]
\centering
\begin{tabular}{|l|l|l|}
\hline
\textbf{Parameter} & \textbf{Datatype} & \textbf{Value} \\ \hline
Tries             & Integer           & 0.99          \\ \hline
Max Step          & Integer           & 2000          \\ \hline
T min             & Integer           & 0             \\ \hline
T max             & Integer           & 100           \\ \hline
Damping           & Float             & 0.0           \\ \hline
Alpha             & Float             & 30            \\ \hline
\end{tabular}
\caption{Simulated Annealing Hyperparameter Configuration}
\label{tab:sa_hayperparameters}
\end{table}

\subsection{Metrics}
The metric for evaluating link configuration on the satellite transponder is the total reward, which reflects the learning behavior of the model. This total reward comprises eight metrics, each linked to specific conditions in the link configuration. The metrics are detailed below along with their calculation methods. They are categorized into Links Reward (LR) and Transponder Reward (TR), contributing 70\% (0.7) and 30\% (0.3) to the total reward, respectively. Individual metrics can be weighted to emphasize more challenging metrics (through the parameters $\omega$, $\theta$, $\phi$, $\mu$, $\beta$, $\epsilon$, $\psi$, $\rho$), potentially enhancing policy learning for the PPO agent. However, the impact of different weightings was not explored in this study.

Links reward metrics include Overlap Reward, On Transponder Reward, PEB Reward, and Margin Reward, which pertain to each link individually. Since multiple links are typically configured on the transponder, the link reward is divided by the number of links to represent the reward per link-weight (RpL). This share, multiplied by the Individual Link Reward for each link, contributes to the total reward (see Equation \ref{individual_link_rew_formula}). Metrics for each link are:

\begin{itemize}
    \item Overlap Reward (OR): Checks if the selected link does not overlap with other links in terms of bandwidth. A reward of 1 is given if met; otherwise, 0.
    \begin{equation}
    \text{OR} = \text{RpL} \times \omega \times or
    \end{equation}
    
    \item On Transponder Reward (OTR): Verifies whether the link's bandwidth is fully on the transponder. If met, the reward is 1; if not, 0.
    \begin{equation}
    \text{OTR} = \text{RpL} \times \theta \times tr
    \end{equation}
    
    \item PEB Reward (PEBR): Assesses the appropriateness of the power-to-bandwidth ratio for the link. A reward of 1 is given if met; otherwise, 0.
    \begin{equation}
    \text{PEBR} = \text{RpL} \times \phi \times pebr
    \end{equation}
    
    \item Margin Reward (MR): Evaluates the EIRP of the link. If it is below the minimum required for high-quality transmission, no reward is given. If exactly at the minimum, the reward is 1. If above, the reward decreases as the EIRP exceeds the minimum.
    \begin{equation}
    \text{MR} = \text{RpL} \times \mu \times mr
    \end{equation}  
\end{itemize}

\begin{equation}
\label{individual_link_rew_formula}
\begin{aligned}
\text{Individual\_Link\_Reward} = \\
\sum (\text{OR} + \text{OTR} + \text{PEBR} + \text{MR})
\end{aligned}
\end{equation}

TR metrics relate to all links or the system as a whole, it includes:

\begin{itemize}
    \item Bandwidth Reward (BR): Measures the ratio of the sum of link bandwidths to the transponder bandwidth. The PPO agent should learn that this sum must not exceed the transponder bandwidth. A reward of 1 is given if met; otherwise, 0.
    \begin{equation}
    \text{BR} = \text{TR} \times \beta
    \end{equation}
    
    \item EIRP Reward (ER) : Assesses the ratio of total link EIRP to transponder EIRP. The PPO agent should learn that the sum of link EIRPs must not exceed the transponder EIRP to avoid overloading. A reward of 1 is given if met; otherwise, 0.
    \begin{equation}
    \text{ER} = \text{TR} \times \epsilon
    \end{equation}

    \item Packed Reward (PR):This metric measures the utilization of overlapping frequency ranges for links on the transponder. A reward is given based on how well the overlapping frequencies are managed, encouraging configurations that maximize frequency usage without excessive interference.
    \begin{equation}
    \text{PR} = \text{TR} \times \psi
    \end{equation}

    \item Free Resource Reward (FRR): This metric evaluates the amount of unused or free resources (e.g., bandwidth or power) on the transponder. It likely rewards configurations that conserve resources, promoting efficient use of the transponder's limited capabilities.
    \begin{equation}
    \text{FRR} = \text{TR} \times \rho
    \end{equation}
    
\end{itemize}

\begin{equation}
\begin{aligned}
\text{Transponder\_Reward} = \\
\sum (\text{BR} + \text{ER} + \text{PR} + \text{FRR})
\end{aligned}
\end{equation}

\subsection{Training Process}
Before initiating the experiments, the learning rate for the PPO was examined through a grid search. 
For the first experiment, the grid search included values of $1e-3$, $5e-4$, $1e-4$, $5e-5$,
and $1e-5$. Each learning rate was evaluated using three different seeds. Across all runs, $2,000,000$ steps were executed in the transponder environment, maintaining a consistent batch size of $2,000$ for comparability. Each episode consisted of $10$ steps, serving as the termination criterion, resulting in a total of $200,000$ episodes. The best performance was obtained with a learning rate of $1e-5$. This learning rate was subsequently used for Experiment 1.

A similar grid search was performed for the second experiment, with learning rate values set at $1e-3$, $1e-4$, and $1e-5$. Each learning rate was again tested with three different seeds. For all runs, $1,000,000$ steps were carried out in the transponder environment, with a batch size of $2,000$. Episodes consisted of $100$ steps. Also for Experiment 2, a learning rate of $1e-5$ yielded the best performance.

\section{RESULTS} \label{sec:results}

\subsection{Experiment 1}
The objective of the first experiment was to determine an optimal link configuration for three links within the transponder environment, ensuring that the bandwidth and EIRP resources of the satellite transponder were conserved. To address this optimization problem, the three algorithms — PPO, Simulated Annealing, and Random Action — were implemented in the transponder environment. The experiment aimed to evaluate whether a trained PPO model could achieve higher performance compared to the Simulated Annealing algorithm. Action Space 1 was used in the environment for this experiment.

\begin{figure}[h!]
    \centering
    \includegraphics[width=\linewidth]{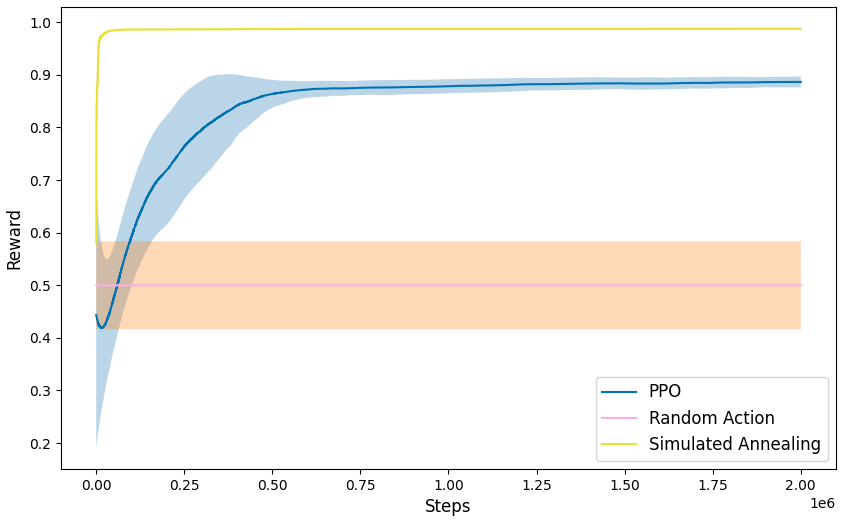}
    \caption{Comparison PPO, Simulated Annealing and Random Action. The x-axis represents the training steps; the y-axis shows the achieved Reward.}
    \label{fig:comparison_plot}
\end{figure}

Figure \ref{fig:comparison_plot} illustrates the training of the PPO model over $2,000,000$ steps, compared to the Simulated Annealing and Random Action baselines. Each PPO episode consisted of $10$ steps. To enhance visualization, all curves were smoothed. Training for the PPO model and the computation of the comparative baselines were conducted with five different seeds ($0$ to $4$). The figure presents the average performance across these runs, including the standard deviation for each algorithm. The maximum achievable value is $1$. It is noteworthy that only the PPO model was trained, as neither the Simulated Annealing algorithm nor the Random Action model can be trained. These baselines were plotted to provide a benchmark for assessing the PPO's performance.

Both the Simulated Annealing and Random Action models were run for 2,000,000 steps. As expected, the Random Action model showed the lowest performance, maintaining an average value of $0.496$ across all steps. The PPO model started at a similar value, which is typical as no policy has been learned at the initial stage. The policy improved continuously over the steps, as indicated by an increase in reward, with most adjustments occurring within the first $500,000$ steps. After this, the PPO curve gradually converged to just below $0.9$, indicating that an effective policy had been learned. The Simulated Annealing algorithm demonstrated the highest performance, rapidly reaching a value just under $1$ within the first $20,000$ steps and converging toward $1$ thereafter.

\begin{figure}[h!]
    \centering
    \includegraphics[width=\linewidth]{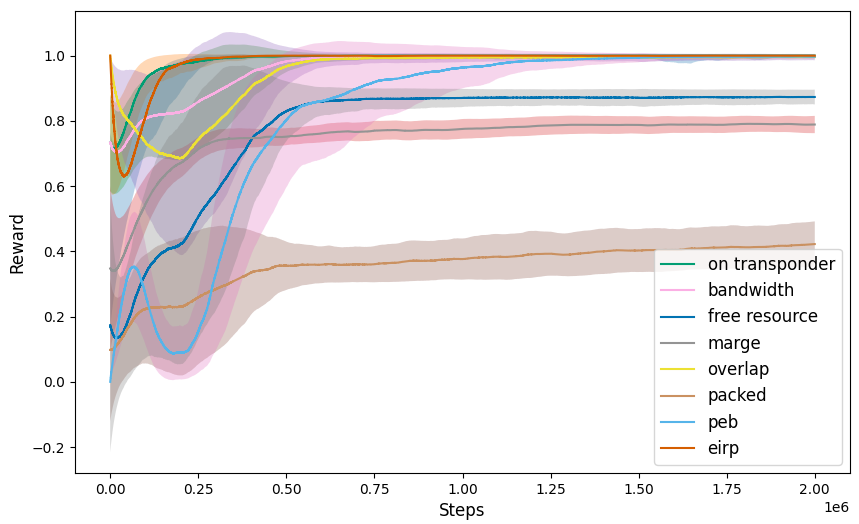}
    \caption{Development of Rewards under the PPO algorithm Training. The learning behavior of the PPO with regards to the various reward components is illustrated. The steps are printed on the x-axis, the reward on the y-axis.}
    \label{fig:comparison_plot_rewards}
\end{figure}

Figure \ref{fig:comparison_plot_rewards} depicts the learning behavior of the PPO model for the individual metrics that collectively contribute to the total reward. These metrics represent conditions that should be met, as explained in Section 5.4. The On Transponder Reward, EIRP Reward, Overlap Reward, PEB Reward, and Bandwidth Reward metrics were effectively learned by the PPO, each reaching the maximum value of 1. A second group of metrics, including Packed Reward, Free Resource Reward, and Margin Reward, showed slower learning progress and had not reached their maximum values by the end of training. The slight upward trend of these curves at the end of training suggests that additional training could further improve the PPO model's performance.

To evaluate the performance of the PPO model finally, the trained PPO model, the Simulated Annealing model, and the Random Action model were applied to unseen observations, referred to as inference. Unlike the training phase, the PPO does not refine its policy during this step but applies the learned policy. Five seeds ($0$–$4$) were used for evaluation. In each run, five different random observations were generated for the configuration of three links. During an episode, the PPO model's task was to configure the links optimally. Each episode consisted of $10$ steps, with the PPO applying its policy learned over $2,000,000$ training steps.

The Random Action model achieved an average value of $0.496$. The Simulated Annealing model reached a value of $0.988$, requiring $2,000,000$ steps to do so. However, as shown in Figure \ref{fig:comparison_plot}, only $20,000$ steps would have been sufficient, as the Simulated Annealing model's performance converged toward $1$ from that point onward. Detailed results of the first experiment are provided in Table \ref{tab:experiment1_results}. The evaluation demonstrates that the Simulated Annealing model achieved the best performance, with a value of $0.988$. The PPO model also performed well, achieving a value of $0.875$. It is noteworthy that the optimization of the PPO model only considered the learning rate through a grid search, while other potentially beneficial hyperparameters were not explored.

\begin{table}[h!]
\centering
\begin{tabular}{|l|c|}
\hline
\textbf{Algorithmus} & \textbf{Reward} \\ \hline
PPO (Action Space 1) & $0.875 \pm 0.007$ \\ \hline
Simulated Annealing  & $\mathbf{0.988 \pm 0.001}$ \\ \hline
Random Action        & $0.496 \pm 0.004$ \\ \hline
\end{tabular}
\caption{Experiment 1: Inference}
\label{tab:experiment1_results}
\end{table}

\subsection{Experiment 2}
Experiment $2$ differs from Experiment $1$ in that it uses Action Space $2$, defined such that a parameter value can be modified from a link in a single step. The parameters available for modification include center frequency, EIRP, and the MOD-FEC combination. As in Experiment 1, PPO was compared with the Simulated Annealing and Random Action baselines. The selected learning rate for PPO was $1e-5$, as this yielded the best performance during the grid search for Action Space $2$.

\begin{figure}[h!]
    \centering
    \includegraphics[width=\linewidth]{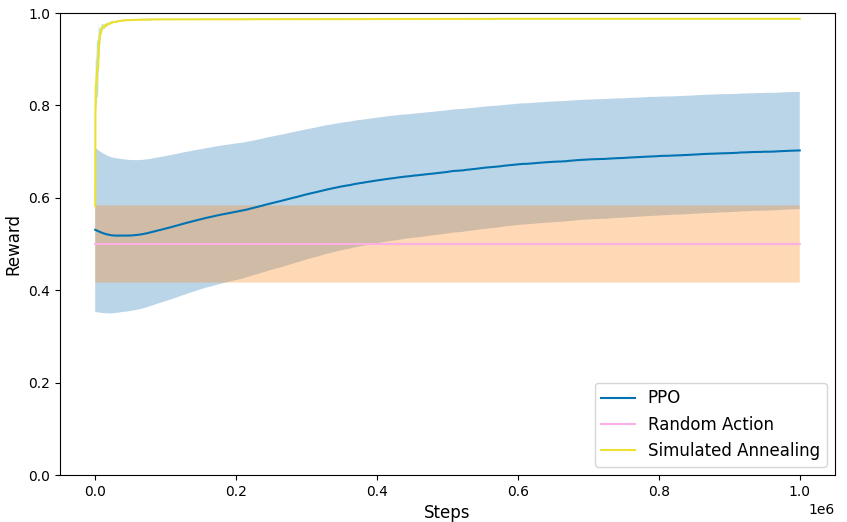}
    \caption{Comparison PPO, Simulated Annealing and Random Action. The x-axis represents the training steps; the y-axis shows the achieved Reward.}
    \label{fig:comparison_plot_as2}
\end{figure}

Figure \ref{fig:comparison_plot_as2} illustrates the training process for Experiment $2$, where the PPO model was trained over $1,000,000$ steps, with each episode comprising $100$ steps. The Simulated Annealing and Random Action baselines were also executed for $1,000,000$ steps. The graphs for Simulated Annealing and Random Action are identical to those in Experiment $1$. As shown, the most significant adjustments in PPO occur during the first $200,000$ training steps.

\begin{figure}[h!]
    \centering
    \includegraphics[width=\linewidth]{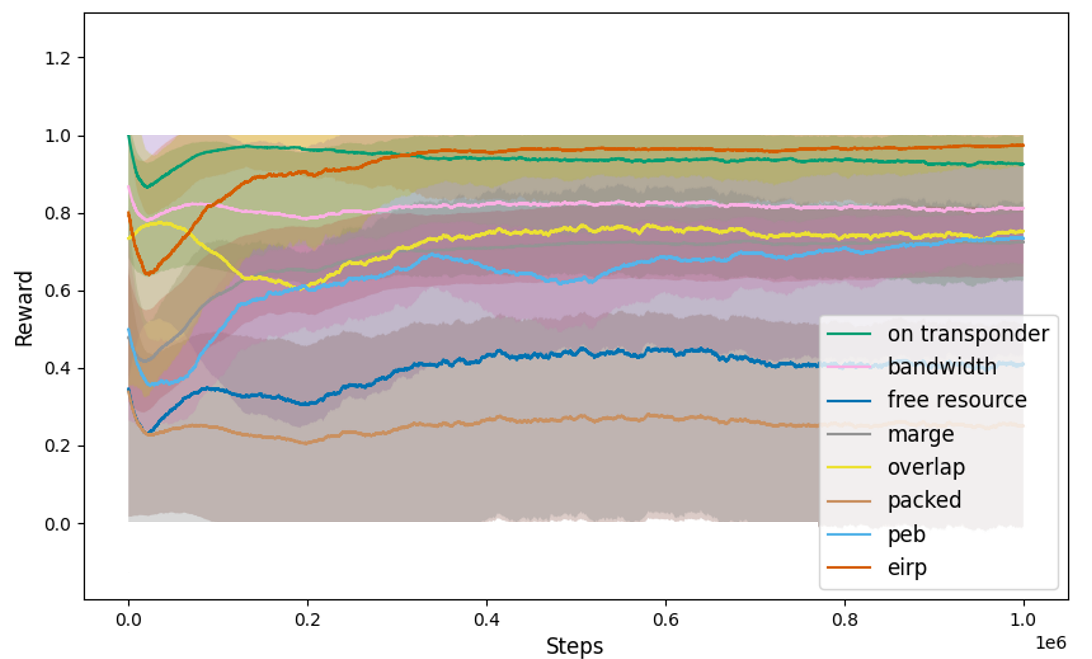}
    \caption{Development of Rewards under the PPO algorithm Training. The learning behavior of the PPO with regards to the various reward components is illustrated. The steps are printed on the x-axis, the reward on the y-axis.}
    \label{fig:comparison_plot_rewards_as2}
\end{figure}

Subsequently, the PPO curve converges toward a value of approximately $0.7$, suggesting that the optimization problem may be somewhat too complex for the defined Action Space $2$. Given that the PPO curve in Figure \ref{fig:comparison_plot_as2} continues to rise slowly, further training could potentially enhance the PPO model's performance. However, the individual metrics contributing to the Total Reward metric are less well-learned with Action Space $2$ compared to Action Space $1$, as indicated in Figure \ref{fig:comparison_plot_rewards_as2}. The EIRP Reward is the only metric fully learned, while the other metrics learn more slowly or converge at different levels after brief training. The Figure suggests that extended training could improve the total reward, although it remains uncertain how much the PPO model can improve, given that a significant number of metrics are not effectively learned by the policy.

Also for the second experiment, the models were evaluated on unseen data. As in Experiment $1$, the Simulated Annealing algorithm achieved the best performance. The PPO model reached a value of $0.786$, performing worse than the Simulated Annealing algorithm. However, since the PPO model's value was notably higher than that of the Random Action model, it can be inferred that the PPO model had effectively learned.

The lower performance of the PPO model compared to Experiment $1$ could be attributed to the action space implementation not sufficiently capturing the complexity of the problem. As discussed, the results of Experiment $2$ should be considered an initial assessment. Further improvements in performance could be achieved by adjusting hyperparameters, modifying the weighting of the action space metrics, or employing reward shaping or curriculum learning.

\begin{table}[h!]
\centering
\begin{tabular}{|l|c|}
\hline
\textbf{Algorithmus} & \textbf{Reward} \\ \hline
PPO (Action Space 2) & $0.786 \pm 0.104$ \\ \hline
Simulated Annealing  & $\mathbf{0.988 \pm 0.001}$ \\ \hline
Random Action        & $0.496 \pm 0.004$ \\ \hline
\end{tabular}
\caption{Experiment 2: Inference}
\label{tab:experiment2_results}
\end{table}

\subsection{Interpretation}
The results achieved with Action Space $1$ are superior to those obtained with Action Space $2$. One possible explanation for this is that, during training, the randomly generated actions in Action Space $1$ evenly scan the action space. In contrast, with Action Space $2$, certain regions are densely sampled while others receive little to no sampling. Consequently, inference using an Action Space $2$ PPO model yields very good results in well-sampled areas and significantly poorer results in sparsely sampled areas.

Another possible reason for the better performance of Action Space $1$ could be the simpler and more direct relationship between the Action Space and Observation Space, making it easier to learn. In Action Space $1$, there is a one-to-one correspondence between the parameters in the Action Space and those in the Observation Space. In Action Space $2$, however, this mapping is more complex, complicating the learning process.

\section{CONCLUSION} \label{sec:conclusion}
Even today, the configuration of connections in satellite communication networks is often performed manually. Given the increasing complexity of the hardware used both in satellites and on the ground, the application of AI in this domain is a logical progression. In this work, a transponder environment was developed for link configuration on a satellite transponder, and the potential utility of RL for solving a static problem was investigated. Two experiments were conducted, differing in their implementation of the action space. In both, a PPO model was compared to two baselines: Simulated Annealing and the Random Action model. The results showed that in both experiments, the Simulated Annealing metaheuristic outperformed the PPO model. Since only the learning rate was optimized for the PPO, further improvements could be achieved through additional hyperparameter tuning. Additionally, the weighting of the metrics was not explored, which could be another factor for enhancing PPO performance.

It is likely that the performance gap between Simulated Annealing and PPO would widen as the static problem becomes more complex and realistic. Simulated Annealing is specifically designed for efficient exploration of large search spaces, potentially demonstrating even greater superiority over PPO in more complex scenarios. Given the growing complexity of this field, continued research is essential to find solutions to this optimization problem. The environment must be refined to better match real-world requirements. For instance, only a single data rate was used in this study, whereas in practice, this parameter is continuous or must be represented in a rasterized form. The mutual interference between links was neglected and needs to be incorporated for practical applications. The restriction to exactly three links must also be removed. An important next step is to transition from a static to a dynamic model, allowing each link to have an active time period. The agent must also evolve alongside the environment, with the neural network topology becoming more complex. A potential strength of PPO could be its ability to handle scenarios where links on the transponder are dynamically added or terminated over time, leveraging its capacity to learn policies that comprehend underlying metrics. However, this potential requires further experimentation and comprehensive research.

Overall, these future steps present an opportunity to enhance the developed model, align it with the complexity of real-world environments, and validate its practical applicability.

\section*{\uppercase{Acknowledgements}}
This paper was partially funded by the German Federal Ministry of Education and Research through the funding program "quantum technologies - from basic research to market" (contract number: 13N16196). Furthermore, this paper was also partially funded by the German Federal Ministry for Economic Affairs and Climate Action through the funding program "Quantum Computing -- Applications for the industry" (contract number: 01MQ22008A).

\end{document}